\documentclass[a4paper]{article}
\usepackage[table]{xcolor}
\usepackage[final]{neurips_2023}
\usepackage{bbding}

\usepackage{natbib}
\setcitestyle{numbers,square}

\usepackage[utf8]{inputenc} 
\usepackage[T1]{fontenc}    
\usepackage{url}            
\usepackage{booktabs}       
\usepackage{amsfonts}       
\usepackage{nicefrac}       
\usepackage{microtype}      
\usepackage{xcolor}         
\usepackage{multirow}
\usepackage{graphicx}
\usepackage{CJKutf8}
\usepackage[many]{tcolorbox}

\usepackage{tabularx}
\usepackage{booktabs}
\usepackage{makecell}
\usepackage[table]{xcolor}
\usepackage{enumitem}
\usepackage{graphicx}  
\usepackage{subcaption}
\usepackage{adjustbox}
\usepackage{fancyvrb}
\usepackage{multirow}
\usepackage{float}
\usepackage{hyperref}  
\usepackage{cleveref}
\usepackage{amsmath}
\usepackage{fancyhdr}

\hypersetup{
  colorlinks=false,    
  citebordercolor=red, 
  linkbordercolor=green,
  pdfborder={0 0 0.8}  
}

\renewenvironment{abstract}{%
  \vskip 0.075in%
  \centerline{\large\bf Abstract}%
  \vskip 0.075in%
  \noindent\ignorespaces%
}{%
  \vskip 0.1in%
}

\definecolor{green}{RGB}{0,150,10}
\definecolor{blue}{RGB}{0,148,181}
\definecolor{orange}{RGB}{194,153,107}

\title{MonkeyOCR v1.5 Technical Report: Unlocking Robust Document Parsing for Complex Patterns}

\author{
Jiarui Zhang\textsuperscript{1},
Yuliang Liu\textsuperscript{2},
Zijun Wu\textsuperscript{1},
Guosheng Pang\textsuperscript{1}, 
Zhili Ye\textsuperscript{1},
Yupei Zhong\textsuperscript{1}, \\
\textbf{Junteng Ma}\textsuperscript{1}, 
\textbf{Tao Wei}\textsuperscript{1}, 
\textbf{Haiyang Xu}\textsuperscript{1}, 
\textbf{Weikai Chen}\textsuperscript{1}, 
\textbf{Zeen Wang}\textsuperscript{1}, 
\textbf{Qiangjun Ji}\textsuperscript{1}, \\
\textbf{Fanxi Zhou}\textsuperscript{1}, 
\textbf{Qi Zhang}\textsuperscript{1}, 
\textbf{Yuanrui Hu}\textsuperscript{1}, 
\textbf{Jiahao Liu}\textsuperscript{1}, 
\textbf{Zhang Li}\textsuperscript{2}, 
\textbf{Ziyang Zhang}\textsuperscript{2}, \\
\textbf{Qiang Liu}\textsuperscript{1}, 
\textbf{Xiang Bai}\textsuperscript{2}
\\
\\
\textsuperscript{1} KingSoft Office Zhuiguang AI Lab, %
\textsuperscript{2} Huazhong University of Science and Technology %
}

\begin{document}

\maketitle

\begin{figure}[!htbp]
\centering
\vspace{-1.1cm}
\includegraphics[width=0.95\linewidth]{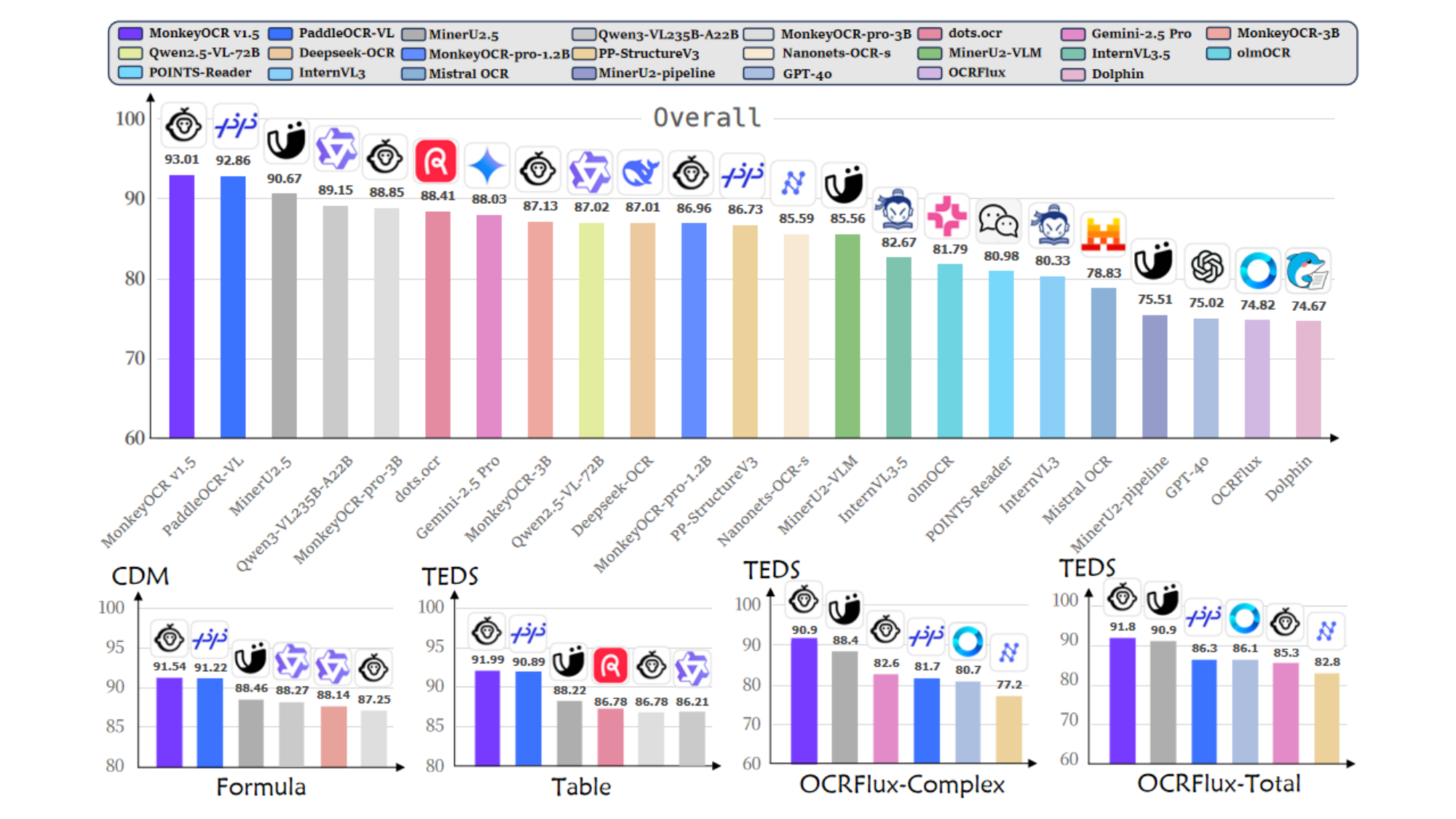}
\caption{Performance comparison of MonkeyOCR v1.5 and other SOTA models.}
\label{fig:abstract}
\vspace{-0.3cm}
\end{figure}

\begin{abstract}
Document parsing is a core task in document intelligence, supporting applications such as information extraction, retrieval-augmented generation, and automated document analysis. However, real-world documents often feature complex layouts with multi-level tables, embedded images or formulas, and cross-page structures, which remain challenging for existing OCR systems. We introduce \textbf{MonkeyOCR v1.5}, a unified vision–language framework that enhances both layout understanding and content recognition through a two-stage pipeline. The first stage employs a large multimodal model to jointly predict layout and reading order, leveraging visual information to ensure sequential consistency. The second stage performs localized recognition of text, formulas, and tables within detected regions, maintaining high visual fidelity while reducing error propagation. To address complex table structures, we propose a \emph{visual consistency–based reinforcement learning} scheme that evaluates recognition quality via render-and-compare alignment, improving structural accuracy without manual annotations. Additionally, two specialized modules, \emph{Image-Decoupled Table Parsing} and \emph{Type-Guided Table Merging}, are introduced to enable reliable parsing of tables containing embedded images and reconstruction of tables crossing pages or columns. Comprehensive experiments on \textbf{OmniDocBench v1.5} demonstrate that MonkeyOCR v1.5 achieves state-of-the-art performance, outperforming PPOCR-VL and MinerU 2.5 while showing exceptional robustness in visually complex document scenarios. A trial link can be found at github link \url{https://github.com/Yuliang-Liu/MonkeyOCR}.
\end{abstract}

\newpage
\tableofcontents
\newpage

\section{Introduction}
\label{sec:intro}

\begin{figure}[b]
\centering
\includegraphics[width=\linewidth]{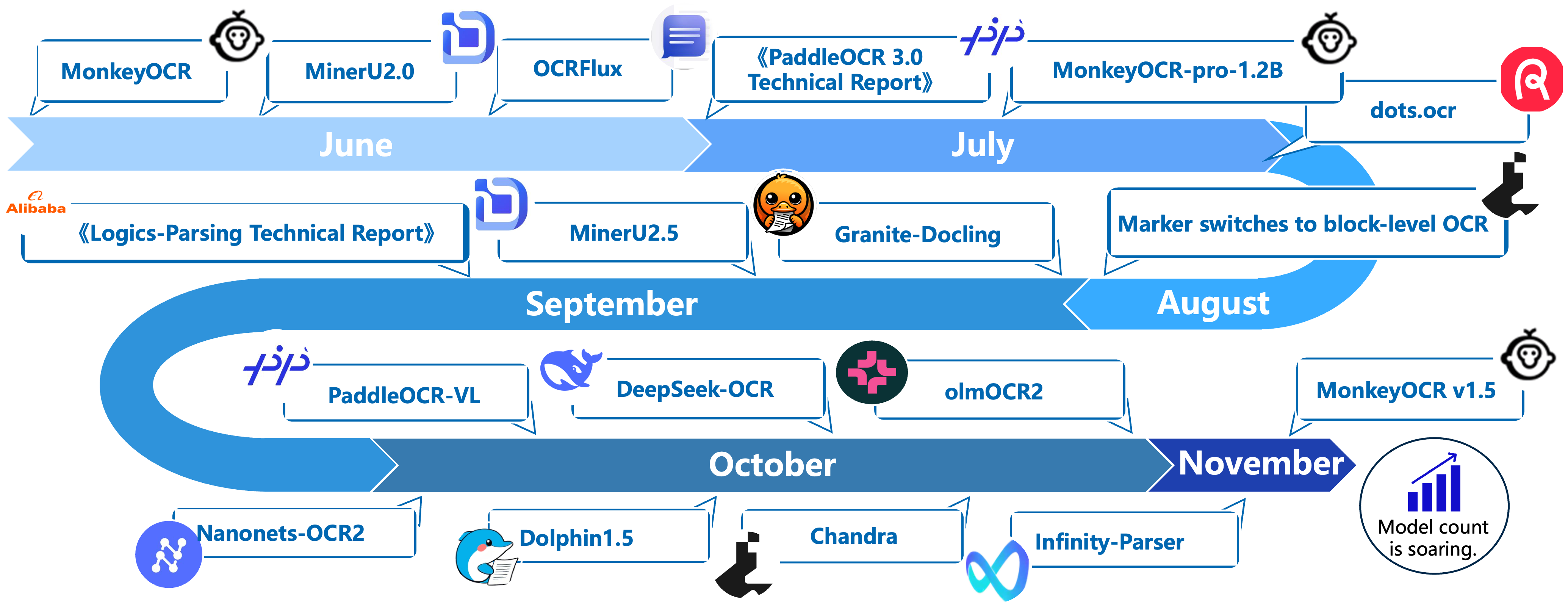}
\caption{Rapid growth of document parsing methods since June 2025.}
\label{fig:overview}
\end{figure}

Document parsing is a fundamental task in the field of document intelligence, serving as the backbone for downstream applications such as information extraction, retrieval-augmented generation, and intelligent document analysis. The goal of document parsing is to systematically transform the complex multimodal contents of various document types, such as scanned images and PDFs, which include text, tables, images, and formulas, into structured representations. However, document images often exhibit highly sophisticated layouts and intricate table structures, posing significant challenges for parsing. Specifically, tables may contain multi-level nesting, cross-page spans, merged or split cells, and embedded elements such as images, formulas, or mixed fonts, all of which complicate accurate content recognition, relational inference, and structured representation. In addition, irregular layouts, diverse languages, and varying typographic styles further increase the demand for robust and generalizable parsing models.

Traditional pipeline-based approaches~\cite{mineru, Cui2025PaddleOCR3T} break down document parsing into a series of subtasks, such as layout detection, text and formula detection, text recognition, and table or formula recognition, with each handled by a dedicated model. This multi-stage process is prone to error accumulation, where mistakes in earlier steps propagate and compromise overall performance. In contrast, end-to-end models~\cite{qwen2.5-vl, internvl, gpt4o, gemini25} process the entire document image in a single pass; however, the high resolution of document images produces a massive number of visual tokens, and the quadratic complexity of self-attention mechanisms creates a significant computational bottleneck. To address these limitations, MonkeyOCR~\cite{li2025monkeyocr} proposed the SRR paradigm, which decouples document parsing into structure detection, content recognition, and reading order prediction. This design streamlines the conventional multi-stage pipeline, effectively mitigating cumulative errors while avoiding the substantial computational overhead associated with full-page end-to-end processing, thereby advancing intelligent multimodal document understanding (Fig.~\ref{fig:overview}). Mineru 2.5~\cite{mineru2.5} further simplify the three-stage framework by employing a unified large multimodal model to jointly predict document layout and reading order, followed by content recognition. PPOCR-VL~\cite{paddleocrvl} adopts a similar three-stage methodology, utilizing lightweight models for structural analysis and reading order prediction, and subsequently applying a large multimodal model for content recognition.

Despite the significant advancements of existing models, they still face challenges in parsing documents with complex scenarios. This paper introduces MonkeyOCR v1.5, a novel document parsing framework that demonstrates superior performance on tasks involving complex layouts and content recognition.
To enhance layout and reading order recognition, we employ a large multimodal model in a two-stage parsing approach. The first stage performs structure detection and relationship prediction, followed by content recognition in the second stage. This design not only streamlines the conventional MonkeyOCR pipeline but also leverages visual-semantic information to improve the model's ability to determine text sequence in intricate typographical layouts.
For complex table recognition, we propose a reinforcement learning algorithm based on visual consistency. This algorithm evaluates the accuracy of recognition results by comparing the visual consistency between the original image and a rendered version. This approach enhances the model's comprehension and parsing capabilities for complex tables without the need for precise manual annotations.
To handle tables containing images, we have designed the Image-Decoupled Table Parsing (IDTP) method. This approach first detects and masks images within a table, then employs the large multimodal model to predict the table's HTML structure while simultaneously generating placeholders for the masked images. During the post-processing stage, these placeholders are replaced with the original images based on a predefined mapping, resulting in a complete and accurate table representation.
Furthermore, to address tables that span across multiple pages or columns, the framework automatically identifies and merges these segments by combining rule-based matching with BERT-based semantic discrimination, thereby reconstructing the complete table structure.

Experimental results show that MonkeyOCR v1.5 achieves state-of-the-art performance on the widely adopted OmniDocBench v1.5 benchmark, surpassing the previous best-performing methods PPOCR-VL and MinerU 2.5 by 0.15\% and 2.34\%, respectively, in overall performance. Notably, MonkeyOCR v1.5 demonstrates stronger robustness in complex scenarios. On newspaper documents characterized by dense text and intricate layouts, it achieves the best performance among all competitors. Moreover, on the OCRFlux-complex dataset, our method significantly outperforms the previous state-of-the-art PPOCR-VL by 8.2\%. Beyond accuracy improvements, MonkeyOCR v1.5 further extends its capabilities over MonkeyOCR, enabling embedded image recovery, cross-page table reconstruction, and multi-column table merging, demonstrating strong potential in complex real-world document scenarios~\cite{ocrbench, ocrbenchv2}.

\section{MonkeyOCR v1.5}
\label{sec:monkeyocr1.5}

We propose MonkeyOCR v1.5, a vision-language-based document parsing framework designed for robust and efficient OCR in complex, real-world documents. Compared with the previous version, v1.5 simplifies the pipeline into two stages and introduces a visual-consistency-based reinforcement learning paradigm that improves table recognition accuracy without relying on large-scale manual annotations. In addition, to address challenges such as embedded images and cross-page or cross-column table merging that other methods struggle with (Tab.~\ref{tab:model_capability}), we incorporate an image-decoupled and type-guided table recognition module. Together, these advancements make MonkeyOCR v1.5 a scalable and high-fidelity OCR system well-suited for heterogeneous document understanding.

    \begin{table*}[ht]
        \centering
        \resizebox{1\textwidth}{!}{
        \begin{tabular}{l|c|ccc|ccc}
        \toprule
        \multirow{2}{*}{\textbf{Models}} & \textbf{Inserted Image} & \multicolumn{3}{c|}{\textbf{Cross-page Table Merging}} & \multicolumn{3}{c}{\textbf{Cross-column Table Merging}} \\
        \cmidrule(lr){3-5} \cmidrule(lr){6-8}
        & \textbf{Det.\&Rec.} & \textbf{RpHdrCont.} & \textbf{NoHdrCont.} & \textbf{SplitCont.} & \textbf{RpHdrCont.} & \textbf{NoHdrCont.} & \textbf{SplitCont.} \\
        \midrule
        MinerU2.5~\cite{mineru2.5}    & \texttimes & \checkmark & \checkmark & \texttimes & \texttimes & \texttimes & \texttimes \\
        MonkeyOCR~\cite{li2025monkeyocr}    & \texttimes & \texttimes & \texttimes & \texttimes & \texttimes & \texttimes & \texttimes \\
        OCRFlux~\cite{OCRFLUX}      & \texttimes & \checkmark & \checkmark & \texttimes & \texttimes & \texttimes & \texttimes \\
        dotsOCR~\cite{dotsocr}     & \texttimes & \texttimes & \texttimes & \texttimes & \texttimes & \texttimes & \texttimes \\
        PaddleOCR-VL~\cite{paddleocrvl} & \checkmark & \texttimes & \texttimes & \texttimes & \texttimes & \texttimes & \texttimes \\
        MonkeyOCR v1.5  & \textbf{\checkmark} & \textbf{\checkmark} & \textbf{\checkmark} & \textbf{\checkmark} & \textbf{\checkmark} & \textbf{\checkmark} & \textbf{\checkmark}  \\
        \bottomrule
        \end{tabular}}
        \caption{Capability Comparison of Document Processing Models. "Ours" exhibits comprehensive capabilities across all evaluated dimensions, including embedded image restoration, cross-page table merging, and cross-column table merging (repeated header, content continuity, and cell splitting). Other models only support partial functionalities, with significant gaps in handling complex table structures and embedded images. RpHdrCont.:continued table with full header repetition, NoHdrCont.: continued table without headers and SplitCont.: continued table with Row-split.}
        \label{tab:model_capability}
    \end{table*}

\begin{figure*}[h!]
    \centering
    \includegraphics[width=0.98\linewidth]{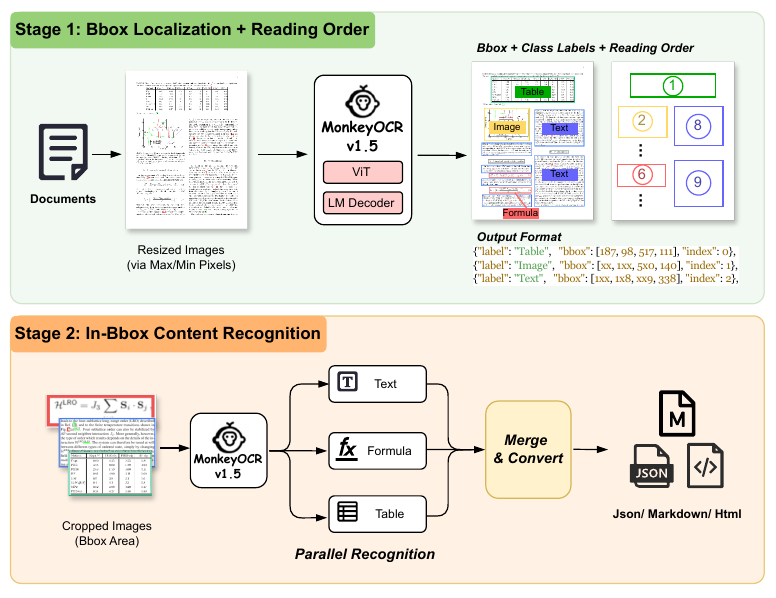}
    \caption{The overall pipeline of MonkeyOCR v1.5, which first detect all layout elements with order index and then recognize contents in a parallel way using a VLM.}
    \label{fig:pipeline}
\end{figure*}

\subsection{Overall Pipeline}
The overall pipeline of \textbf{MonkeyOCR v1.5} consists of two sequential yet lightweight stages: \emph{layout analysis} and \emph{content recognition}, as described in Fig.~\ref{fig:pipeline}. This design leverages the model’s semantic understanding capabilities, enabling more accurate layout detection and reading order prediction.

\textbf{Stage I: Layout detection and reading order prediction.}
In earlier versions, MonkeyOCR relied on text-based models to infer reading order, which often failed to leverage global visual context. To address this, v1.5 employs a vision–language model (VLM) that jointly predicts the page layout and reading order, ensuring stronger visual consistency between detected regions and their spatial order. Given a document image \(I \in \mathbb{R}^{H \times W \times 3}\) and a layout prompt \(p_{\text{layout}}\), the model generates a structured token sequence \(y = \{y_t\}_{t=1}^{T}\) that encodes bounding boxes, indices, categories, and rotation angles:
\begin{equation}
p_{\theta}(y \mid I, p_{\text{layout}}) = \prod_{t=1}^{T} p_{\theta}(y_t \mid y_{<t}, I, p_{\text{layout}}).
\end{equation}
The output follows a constrained JSON schema:
\[
\{\texttt{bbox}:(x_1,y_1,x_2,y_2),\; \texttt{index}:i,\; \texttt{label}:c,\; \texttt{rotation}:\alpha_i\},
\]
where \( (x_1, y_1, x_2, y_2) \) defines the region coordinates, \(i\) is the element index in reading order, \(c\) denotes the category (e.g., text, formula, table), and \(\alpha_i\) represents rotation in degrees. Constrained decoding ensures syntactic validity and geometric consistency while enabling the VLM to infer both structure and order directly from visual cues.

\textbf{Stage II: Region-level content recognition.}
Each detected region \(y_i = (\texttt{bbox}_i, \texttt{label}_i, \alpha_i)\) is cropped from the page and rotated according to its predicted angle to restore upright orientation:
\begin{equation}
I_i = \text{Rotate}\big(\text{Crop}(I, \texttt{bbox}_i), \alpha_i\big).
\end{equation}
The aligned patch \(I_i\) is then passed to the same VLM for semantic decoding, conditioned on the region type:
\begin{equation}
\hat{c}_i =
\begin{cases}
\mathcal{R}_{\text{text}}(I_i), & \text{if } \texttt{label}_i = \text{text},\\[3pt]
\mathcal{R}_{\text{formula}}(I_i), & \text{if } \texttt{label}_i = \text{formula},\\[3pt]
\mathcal{R}_{\text{table}}(I_i), & \text{if } \texttt{label}_i = \text{tablebody}.
\end{cases}
\end{equation}
All recognized elements are finally aggregated according to the predicted reading order \( \pi = \{i_1, i_2, \dots, i_N\} \) to reconstruct the full document representation:
\begin{equation}
\hat{Y}_{\text{doc}} = \text{Merge}\big(\{\hat{c}_{\pi(i)}\}_{i=1}^N\big).
\end{equation}

This two-stage design allows MonkeyOCR v1.5 to efficiently combine global visual reasoning with localized recognition. By coupling layout and reading-order prediction within a single VLM, the system achieves stronger visual–structural consistency and more accurate downstream text, formula, and table recognition compared with modular baselines.

\begin{figure}[h!]
    \centering
    \includegraphics[width=1.0\textwidth]{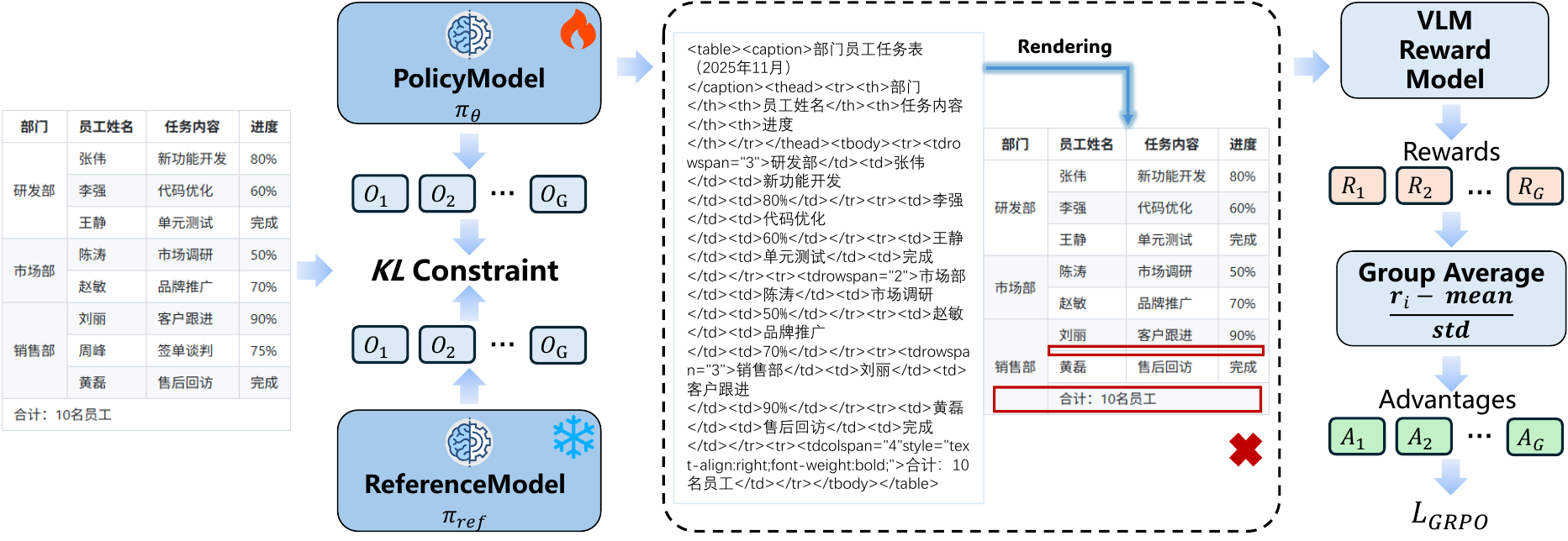}
    \caption{Visual consistency based GRPO. For each input \(x\) containing the original image \(I^{\mathcal{O}}\), the policy model generates a response \(y\). A renderer produces \(I^{\mathcal{R}}\). The triplet \((I^{\mathcal{O}}, y, I^{\mathcal{R}})\) is evaluated by a composite reward that combines a rule-based check with a VLM reward model.}
    \label{fig:grpo}
\end{figure}

\subsection{Visual Consistency-based Reinforcement Learning}

To fully leverage unlabeled data and enhance the model’s capability in recognizing complex tables, we introduce a \textbf{visual consistency-based reinforcement learning} paradigm.  

We first train a \textbf{reward model} based on a vision-language model (VLM). Using the available labeled data, we construct positive–negative sample pairs by modifying the ground-truth (GT) annotations to generate visually inconsistent variants. In addition, we perform multiple samplings with the fine-tuned VLM to produce table recognition outputs; incorrect results are paired with GTs to form additional positive–negative pairs. This strategy enables the reward model to capture typical error patterns and learn fine-grained visual consistency between predicted and reference tables. During training, candidates are scored by the reward model that consumes the triplet $\big(I^{\mathcal{O}}, y, I^{\mathcal{R}}\big)$ and predicts whether $y$ reconstructs the table correctly:

$$
reward = VLM(I^{\mathcal{O}}, y, I^{\mathcal{R}})
$$

After obtaining the trained reward model, we apply the GRPO (Generalized Reinforcement Policy Optimization) algorithm to further optimize the SFT (Supervised Fine-Tuning) model, as illustrated in Fig.~\ref{fig:grpo}. The reward signals are provided by the learned reward model, guiding the policy toward visually consistent outputs.

where $\pi_\theta$ denotes the policy model, $r_\phi(x, y)$ is the reward predicted by the reward model, and $\mathcal{D}$ represents the data distribution (including unlabeled samples).  

This visual consistency-driven reinforcement learning framework enables the model to exploit large-scale unlabeled data, effectively improving table fidelity and robustness without requiring additional manual annotations.

\subsection{Image-Decoupled Table Parsing}

Tables in real-world documents often contain embedded figures, which can confuse OCR systems that treat tables as purely textual images. To address this issue, we propose an \textbf{image-decoupled table parsing} pipeline that separates visual element localization from textual parsing, ensuring that non-textual content does not interfere with cell recognition.

\begin{figure*}[h!]
    \centering
    \includegraphics[width=0.98\linewidth]{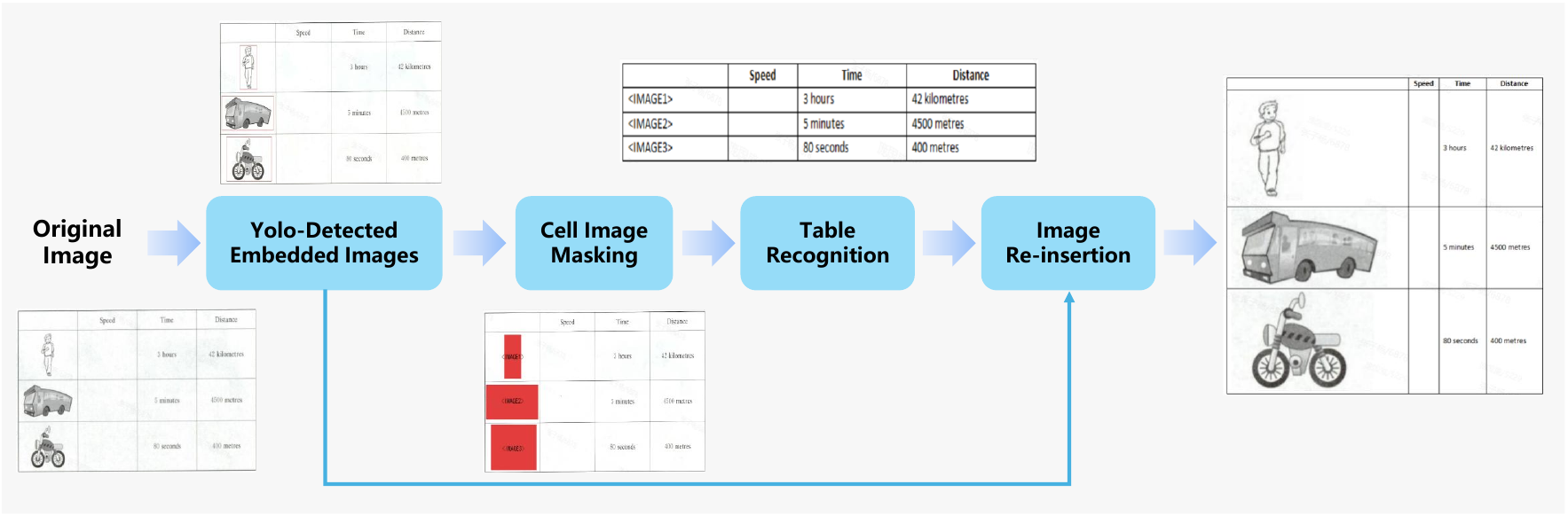}
    \caption{\textbf{Pipeline for tables with embedded images.} The pipeline detects embedded images, replaces them with size-accurate placeholders, performs recognition to generate HTML with \texttt{<img>} tags, and finally re-inserts the original images to reconstruct the table.}
    \label{fig:table_embed}
\end{figure*}

As illustrated in Fig.~\ref{fig:table_embed}, we first employ YOLOv10~\cite{yolov10} to detect embedded images within table regions. Each detected figure is replaced by a precisely sized placeholder mask, while a one-to-one mapping between placeholder IDs and cropped image files is maintained. The masked table is then passed to the recognizer, which is trained with an auxiliary objective encouraging it to treat placeholders as atomic tokens and to produce an intermediate HTML representation containing \texttt{<img>} tags at the corresponding locations.  
During post-processing, each \texttt{<img>} tag is deterministically replaced with its associated image according to the stored mapping, yielding a visually complete table while preserving clean textual outputs. This decoupled design effectively separates image restoration from structural recognition, improving both text accuracy and visual fidelity in complex tables.

\begin{figure}[htbp]
  \centering
  \includegraphics[width=0.9\linewidth]{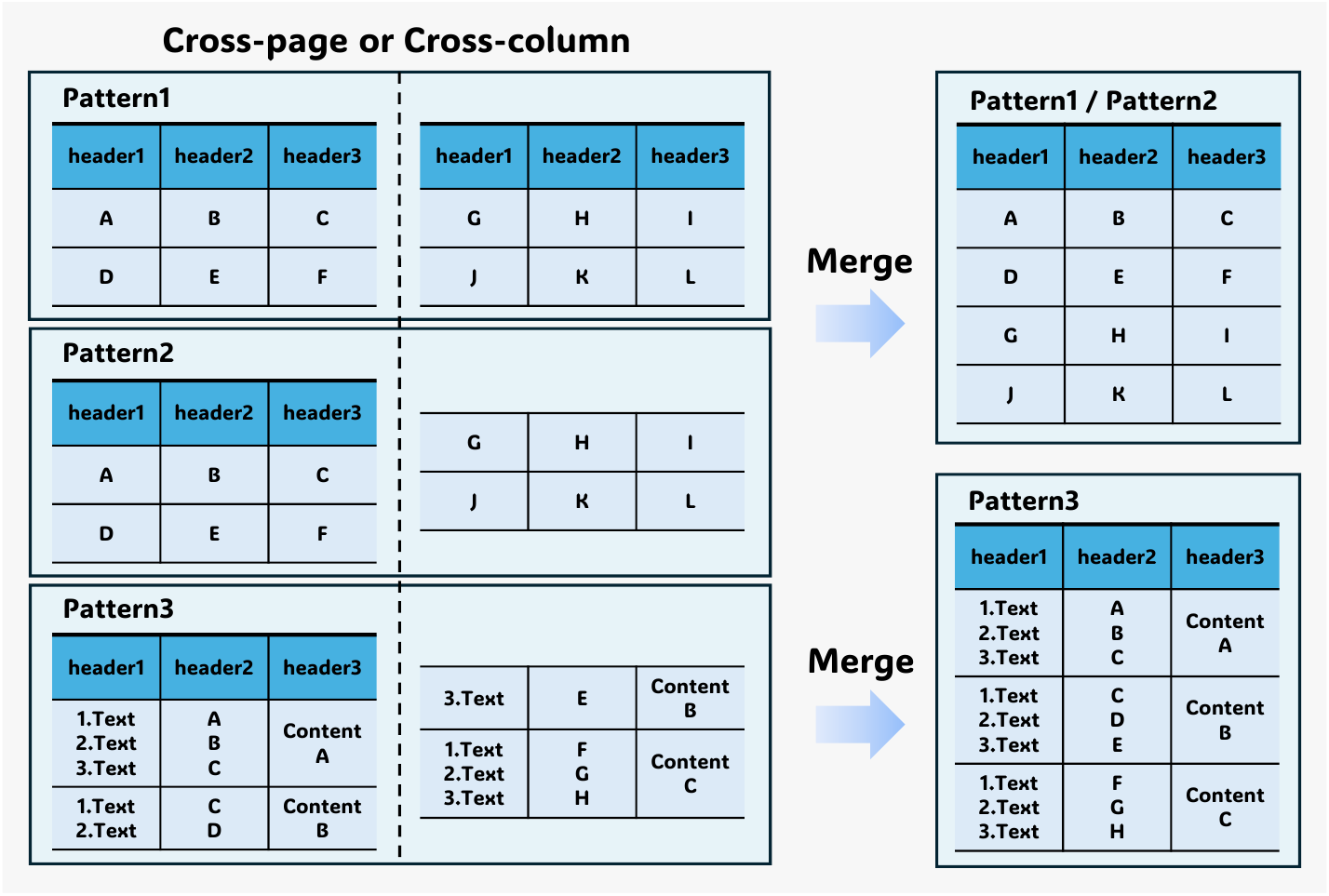}
  \caption{\textbf{Three cross-page/cross-column patterns.} Pattern 1: full header duplication. Pattern 2: continued table without headers. Pattern 3: row-split continuation. }
  \label{fig:table_merge}
\end{figure}

\subsection{Type-Guided Table Merging}

In practice, long tables are often split across pages or columns due to layout constraints. We propose a systematic merging strategy to reconstruct a single coherent table from such fragments, targeting three common patterns (Fig.~\ref{fig:table_merge}):

\begin{itemize}
    \item \textbf{Pattern 1: Full header duplication.}
    If the first rows of adjacent fragments are identical, the second fragment is treated as a continuation with repeated headers. The duplicated header is removed, and the table bodies are concatenated while preserving column alignment.
    \item \textbf{Pattern 2: Continued table without headers.}  
    If the first rows differ but the fragments belong to the same logical table and no cells are split at the boundary, the fragments are directly concatenated, maintaining the existing column schema.
    \item \textbf{Pattern 3: Row-split continuation.}  
    If a cell is split across the fragment boundary, the corresponding spans in both fragments are identified and merged before concatenating the tables, restoring row integrity.
\end{itemize}

\begin{table*}[t]
    \centering
    \resizebox{1\textwidth}{!}{
    \begin{tabular}{c|l|c|ccccc}
    \toprule
    \textbf{Model Type} & \textbf{Methods} & \textbf{Overall↑} & \textbf{Text}\textsuperscript{\textbf{Edit}}↓ & \textbf{Formula}\textsuperscript{\textbf{CDM}}↑ & \textbf{Table}\textsuperscript{\textbf{TEDS}}↑ & \textbf{Table}\textsuperscript{\textbf{TEDS-s}}↑ & \textbf{Read Order}\textsuperscript{\textbf{Edit}}↓ \\
    \midrule
    \multirow{13}{*}{\parbox{1.2cm}{\centering \textbf{Expert}\\\textbf{VLMs}}} 
    & PaddleOCR-VL ~\cite{paddleocrvl}    & 92.86& \textbf{0.035}& 91.22& 90.89& 94.76& 0.043\\
    & MinerU2.5  ~\cite{mineru2.5}      & 90.67& 0.047 & 88.46 & 88.22 & 92.38 & \textbf{0.044} \\
    & MonkeyOCR-pro-3B ~\cite{li2025monkeyocr}    & 88.85& 0.075 & 87.25 & 86.78 & 90.63 & 0.128 \\
    & dots.ocr ~\cite{dotsocr}          & 88.41& 0.048 & 83.22 & 86.78 & 90.62 & 0.053 \\
    & Deepseek-OCR~\cite{deepseekocr}       & 87.01& 0.073 & 83.37 & 84.97 & 88.80 & 0.086\\
    & MonkeyOCR-3B        & 87.13& 0.075 & 87.45 & 81.39 & 85.92 & 0.129 \\
    & MonkeyOCR-pro-1.2B  & 86.96& 0.084 & 85.02 & 84.24 & 89.02 & 0.130 \\ 
    & Nanonets-OCR-s ~\cite{Nanonets-OCR-s}      & 85.59& 0.093 & 85.90 & 80.14 & 85.57 & 0.108 \\
    & MinerU2-VLM ~\cite{2024mineru}         & 85.56& 0.078 & 80.95 & 83.54 & 87.66 & 0.086 \\
    & olmOCR ~\cite{olmocr}              & 81.79& 0.096 & 86.04 & 68.92 & 74.77 & 0.121 \\
    & POINTS-Reader ~\cite{points-reader}       & 80.98& 0.134 & 79.20 & 77.13 & 81.66 & 0.145 \\
    & Mistral OCR ~\cite{mistralocr}         & 78.83& 0.164 & 82.84 & 70.03 & 78.04 & 0.144 \\
    & OCRFlux  ~\cite{OCRFLUX}            & 74.82& 0.193 & 68.03 & 75.75 & 80.23 & 0.202 \\
    & Dolphin  ~\cite{Feng2025DolphinDI}            & 74.67& 0.125 & 67.85 & 68.70 & 77.77 & 0.124 \\
    \midrule
    \multirow{6}{*}{\parbox{1.2cm}{\centering \textbf{General}\\\textbf{VLMs}}} 
    & Qwen3-VL-235B ~\cite{qwen2.5-vl}  & 89.15& 0.069 & 88.14 & 86.21 & 90.55 & 0.068 \\
    & Gemini-2.5 Pro ~\cite{gemini25}  & 88.03& 0.075 & 85.82 & 85.71 & 90.29 & 0.097 \\
    & Qwen2.5-VL-72B  ~\cite{qwen2.5-vl}     & 87.02& 0.094 & 88.27 & 82.15 & 86.22 & 0.102 \\
    & InternVL3.5-241B ~\cite{internvl}     & 82.67& 0.142 & 87.23 & 75.00 & 81.28 & 0.125 \\
    & InternVL3-78B       & 80.33& 0.131 & 83.42 & 70.64 & 77.74 & 0.113 \\
    & GPT-4o ~\cite{gpt4o}          & 75.02& 0.217 & 79.70 & 67.07 & 76.09 & 0.148 \\
    \midrule
    \multirow{3}{*}{\parbox{1.2cm}{\centering \textbf{Pipeline}\\\textbf{Tools}}} 
    & PP-StructureV3   ~\cite{Cui2025PaddleOCR3T}    & 86.73& 0.073 & 85.79 & 81.68 & 89.48 & 0.073 \\
    & Mineru2-pipeline    & 75.51& 0.209 & 76.55 & 70.90 & 79.11 & 0.225 \\
    & Marker-1.8.2 ~\cite{marker}        & 71.30& 0.206 & 76.66 & 57.88 & 71.17 & 0.250 \\

    \midrule
    \textbf{Ours} 
    & MonkeyOCR v1.5      & \textbf{93.01} & 0.045 & \textbf{91.54} & \textbf{91.99} & \textbf{95.04} & 0.049 \\
    \bottomrule
    \end{tabular}}
    \caption{Evaluation on \textsc{OmniDocBench v1.5}.}
    \label{tab:model_evaluation_overall}
    \end{table*}

To operationalize these rules, we adopt a hybrid decision process. Pattern 1 is detected via rule-based header matching with exact or near-exact comparison. To distinguish Pattern 2 from Pattern 3, we employ a BERT-based classifier that predicts whether the leading row of the subsequent fragment semantically continues the trailing row of the preceding fragment. A positive continuation triggers row-level cell merging (Pattern 3), while a negative result leads to concatenation as a headerless continuation (Pattern 2). During merging, we align column schemas, resolve span conflicts, and normalize header tokens to ensure structural consistency. This procedure reconstructs split tables into a single, semantically coherent structure while preserving header continuity and row integrity.

\section{Experiments}
To evaluate our model’s capabilities in general document parsing and complex table recognition, we assess MonkeyOCR v1.5 on OmniDocBench~\cite{omnidocbench} for overall performance, and further evaluate its table recognition on PubTabNet~\cite{pubtabnet} and OCRFlux-pubtabnet-single~\cite{OCRFLUX}. As shown in Fig.~\ref{fig:abstract}, MonkeyOCR v1.5 demonstrates state-of-the-art performance in general document parsing and significantly outperforms the previous best model, PPOCR-VL, by 9.2\% on complex table recognition tasks.

\subsection{Comparison with Other Methods on Different Tasks}

Document parsing involves multiple subtasks, including text recognition, formula recognition, table recognition, and reading-order prediction. Our model achieves the overall best performance, surpassing the previous state-of-the-art expert document parsing model, PPOCR-VL, by 0.15\% on the overall metric, with improvements of 0.32\% in formula recognition and 1.01\% in table recognition. It also outperforms MinerU2.5 by 2.34\% overall, including 3.08\% in formula recognition and 3.77\% in table recognition. Compared with pipeline-based approaches, our model exceeds the best pipeline method, PP-Structure v3, by 6.38\%, demonstrating the effectiveness of our two-stage VLM-based paradigm. Against large general end-to-end models, our method outperforms Qwen-3-VL-235B by 3.86\% and the best closed-source model, Gemini 2.5-Pro, by 4.98\%, highlighting the advantage of specialized models in vertical domains.

\begin{table*}[ht]
    \centering
    \resizebox{1\textwidth}{!}{
    \begin{tabular}{l|l|ccccccccc}
    \toprule
    \textbf{\makecell{Model\\Type}} & \textbf{Models} & \textbf{Slides} & \makecell{\textbf{Academic}\\\textbf{Papers}} & \textbf{Book} & \textbf{Textbook} & \textbf{Exam Papers} & \textbf{Magazine} & \textbf{Newspaper} & \textbf{Notes} & \makecell{\textbf{Financial}\\\textbf{Report}} \\
    \midrule
    \multirow{3}{*}{\makecell{Pipeline\\Tools}}
    & Marker-1.8.2~ ~\cite{marker} & 0.180 & 0.041 & 0.101 & 0.291 & 0.296 & 0.111 & 0.272 & 0.466 & 0.034 \\
    & MinerU2-pipeline~ ~\cite{2024mineru} & 0.424 & 0.023 & 0.263 & 0.122 & 0.082 & 0.395 & 0.074 & 0.260 & 0.041 \\
    & PP-StructureV3~ ~\cite{Cui2025PaddleOCR3T} & 0.079 & 0.024 & 0.042 & 0.111 & 0.095 & 0.072 & 0.062 & 0.124 & 0.018 \\
    \midrule
    \multirow{5}{*}{\makecell{General\\VLMs}} 
    & GPT-4o~ ~\cite{gpt4o} & 0.102 & 0.120 & 0.129 & 0.160 & 0.194 & 0.142 & 0.625 & 0.261 & 0.334 \\
    & InternVL3-76B~ ~\cite{internvl} & 0.035 & 0.105 & 0.063 & 0.083 & 0.101 & 0.041 & 0.583 & 0.092 & 0.067 \\
    & InternVL3.5-241B & 0.048 & 0.086 & \textbf{0.024} & 0.106 & 0.099 & 0.058 & 0.640 & 0.136 & 0.112 \\
    & Qwen2.5-VL-72B ~\cite{qwen2.5-vl} & 0.042 & 0.080 & 0.059 & 0.115 & 0.068 & 0.096 & 0.238 & 0.123 & 0.026 \\
    & Gemini-2.5 Pro~ ~\cite{gemini25} & 0.033 & \underline{0.018} & 0.069 & 0.162 & 0.094 & \textbf{0.016} & 0.135 & 0.117 & 0.017 \\
    \midrule
    \multirow{11}{*}{\makecell{Specialized\\VLMs}}
    & Dolphin~ ~\cite{Feng2025DolphinDI} & 0.096 & 0.045 & 0.062 & 0.133 & 0.168 & 0.070 & 0.239 & 0.256 & 0.019 \\
    & OCRFlux~ ~\cite{OCRFLUX} & 0.087 & 0.087 & 0.082 & 0.184 & 0.207 & 0.105 & 0.730 & 0.157 & 0.019 \\
    & Mistral-OCR~ ~\cite{mistralocr} & 0.092 & 0.053 & 0.061 & 0.135 & 0.134 & 0.058 & 0.564 & 0.310 & 0.052 \\
    & POINTS-Reader~ ~\cite{points-reader} & 0.033 & 0.078 & 0.067 & 0.137 & 0.190 & 0.134 & 0.379 & 0.094 & 0.095 \\
    & olmOCR-7B~ ~\cite{olmocr} & 0.050 & 0.037 & 0.054 & 0.120 & 0.073 & 0.070 & 0.292 & 0.122 & 0.046 \\
    & MinerU2-VLM & 0.075 & \textbf{0.010} & 0.036 & 0.128 & 0.070 & 0.065 & 0.183 & \underline{0.080} & 0.024 \\
    & Nanonets-OCR-s~ ~\cite{Nanonets-OCR-s} & 0.055 & 0.058 & 0.061 & 0.093 & 0.083 & 0.092 & 0.197 & 0.161 & 0.040 \\
    & MonkeyOCR pro-1.2B~ ~\cite{li2025monkeyocr} & 0.096 & 0.035 & 0.053 & 0.111 & 0.089 & 0.049 & 0.100 & 0.169 & 0.020 \\
    & MonkeyOCR pro-3B & 0.090 & 0.036 & 0.049 & 0.107 & 0.075 & 0.048 & 0.096 & 0.117 & 0.020 \\
    & dots.ocr~ ~\cite{dotsocr} & \textbf{0.029} & 0.023 & 0.043 & {0.079} & \textbf{0.047} & \underline{0.022} & 0.067 & 0.112 & \textbf{0.008} \\
    & MinerU2.5 ~\cite{mineru2.5} & \textbf{0.029} & 0.024 & \underline{0.033} & \textbf{0.050} & 0.068 & 0.032 & \underline{0.054} & 0.116 & \underline{0.010} \\
    \midrule
    \textbf{Ours} & MonkeyOCR v1.5 & \underline{0.034} & 0.029 & 0.035 & \underline{0.071} & \underline{0.050} & 0.032 & \textbf{0.049} & \textbf{0.059} & 0.023 \\
    \bottomrule
    \end{tabular}}
    \caption{Results across document types on \textsc{OmniDocBench} (\textbf{bold} = best, \underline{underline} = second best; lower is better).}
    \label{tab:document_type_evaluation}
\end{table*}

\subsection{Comparison with Other Methods on Different Document Types}

As shown in Tab.~\ref{tab:document_type_evaluation},we compare MonkeyOCR v1.5 with other methods on the OmniDocBench benchmark across nine document categories. MonkeyOCR v1.5 achieves performance comparable to existing models on most categories and attains the best results on Newspaper and Notes. The complex layout of Newspaper documents further demonstrates the superiority of our approach in handling challenging real-world scenarios.

\begin{table*}[ht]
\centering
\hfill
\centering
\resizebox{\textwidth}{!}{
\begin{tabular}{l|cccccc}
\toprule
\multirow{1}{*}{\textbf{Dataset}} & Nanonets-OCR~\cite{Nanonets-OCR-s} & MonkeyOCR ~\cite{li2025monkeyocr}& OCRFlux~\cite{OCRFLUX} & MinerU2.5~\cite{mineru2.5} & PaddleOCR-VL~\cite{paddleocrvl}  & Ours \\
\midrule
\textbf{PubTabNet} & - & - & - & 89.1 & 85.2 & \textbf{90.7} \\
\textbf{OCRFlux-Simple} & 88.2 & 88.0 & 91.2 & \textbf{93.3} &   90.7 &92.6 \\ 
\textbf{OCRFlux-Complex} & 77.2 & 82.6 & 80.7 & 88.4 & 81.7&\textbf{90.9} \\
\textbf{OCRFlux-Total} & 82.8 & 85.3 & 86.1 & 90.9&86.3&\textbf{91.8} \\
\bottomrule
\end{tabular}}
\caption{Comparative evaluation of table recognition methods.}
\label{tab:combined_tables}
\end{table*}

\subsection{Comparison with Other Methods on Table Recognition}

To evaluate the performance of our model on table recognition tasks, we conduct experiments on PubTabNet and OCRFlux-pubtabnet-single, comparing MonkeyOCR v1.5 with several state-of-the-art methods, including MinerU2.5 and PaddleOCR-VL. As shown in Table~5, MonkeyOCR v1.5 achieves the best performance on both datasets.
Notably, on the OCRFlux-complex subset, our model surpasses PaddleOCR-VL by 9.2\%, demonstrating the effectiveness of our visual consistency-based reinforcement learning strategy in handling complex table layouts. These results confirm that MonkeyOCR v1.5 not only delivers superior recognition accuracy but also exhibits strong generalization to diverse and structurally intricate table formats commonly found in real-world documents.

\section{Visualization Comparison with Other Methods}

\begin{figure}[H]
\centering
\includegraphics[width=0.95\linewidth]{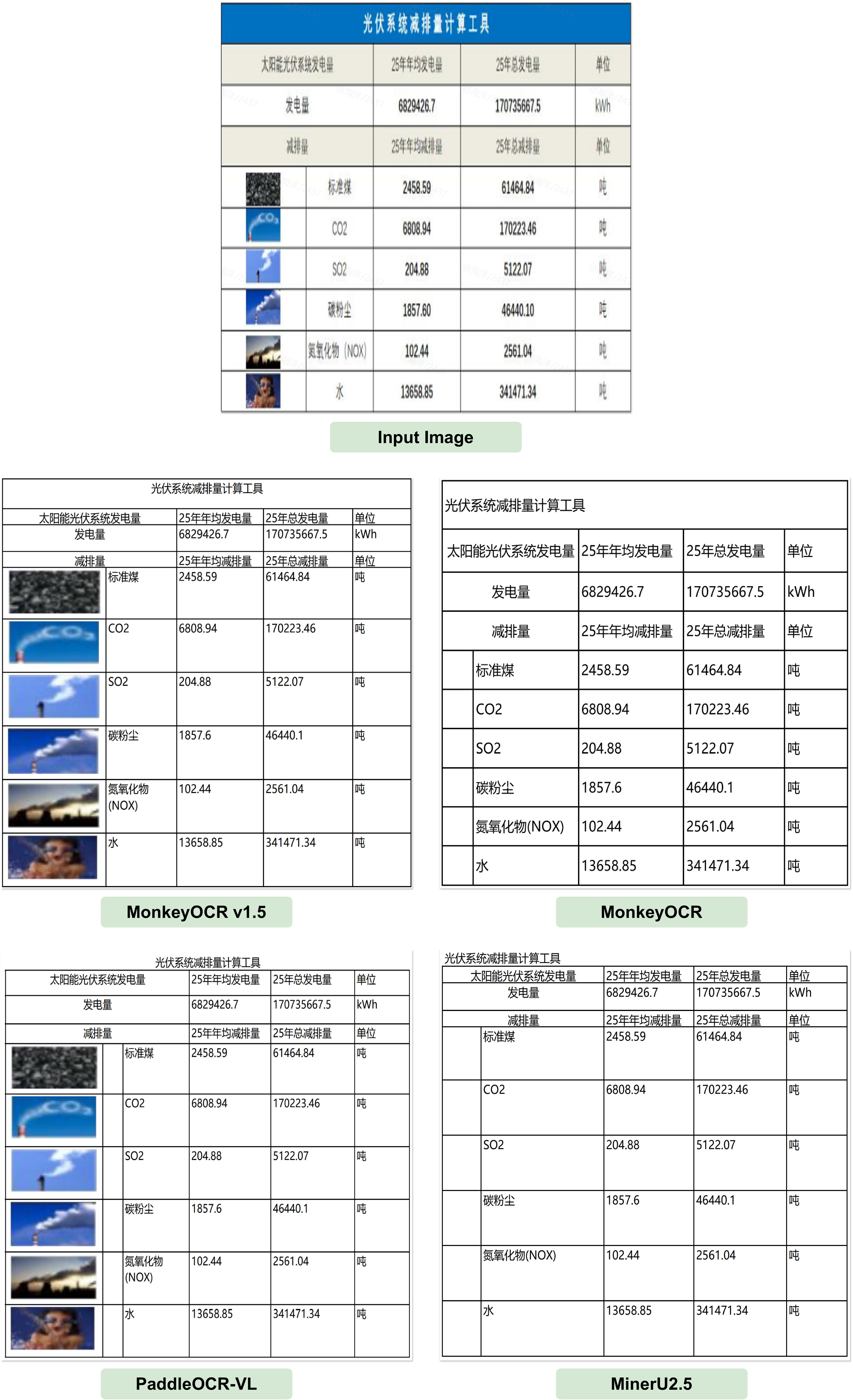}
\caption{\textbf{Embedded Image Detection and Restoration (Example 1)}. Our method perfectly restores the table and its embedded images. In contrast, PaddleOCR-VL falsely detects an extra empty column and loses header cells, while MinerU2.5 fails to restore the images.}
\label{fig:Visualization-2}
\end{figure}

\begin{figure}[H]
\centering
\includegraphics[width=0.95\linewidth]{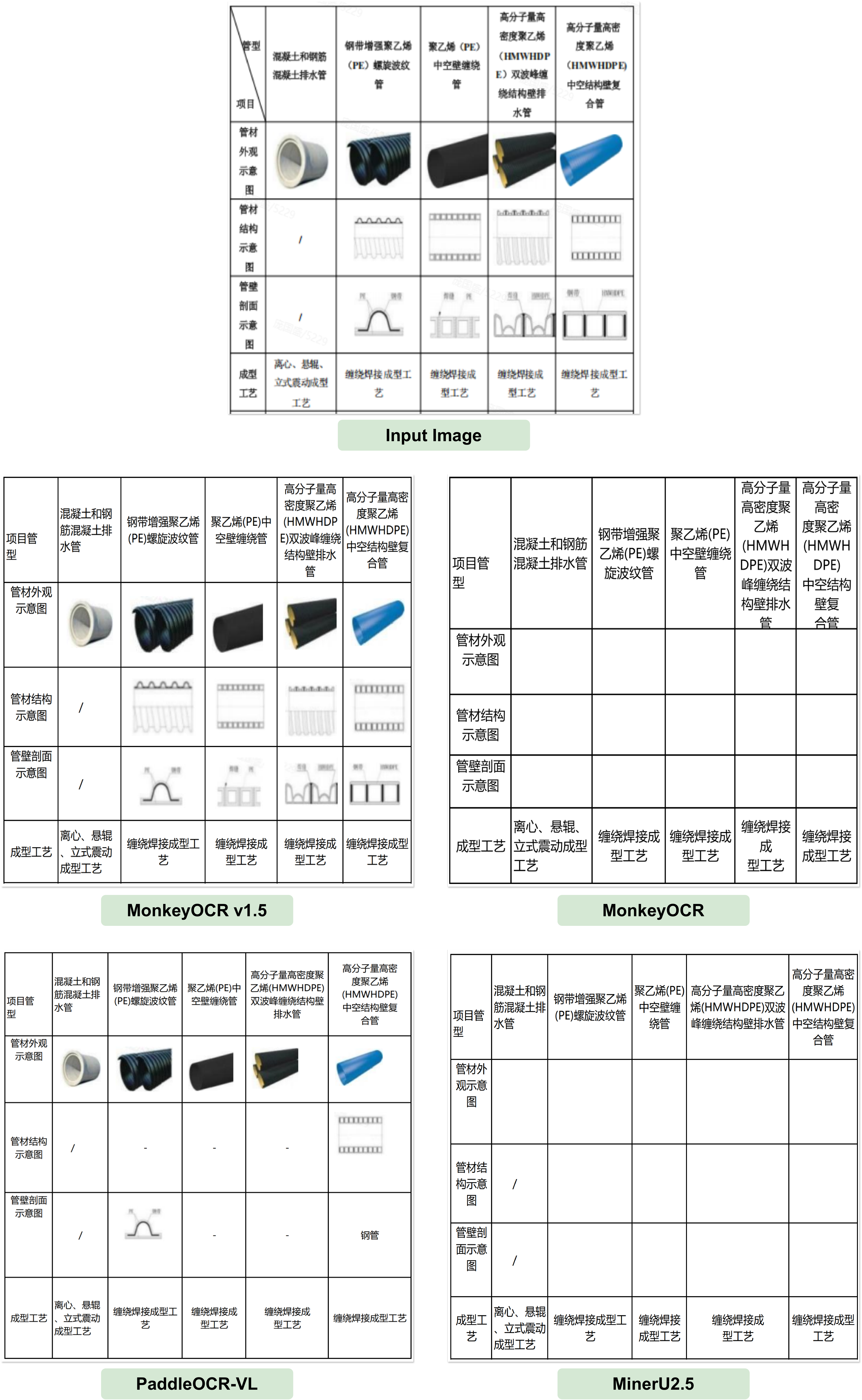}
\caption{\textbf{Embedded Image Detection and Restoration (Example 2)}. Our method restores all images, whereas both PaddleOCR-VL and MinerU2.5 suffer from significant image loss.}
\label{fig:Visualization-3}
\end{figure}

\begin{figure}[H]
\centering
\includegraphics[width=\linewidth]{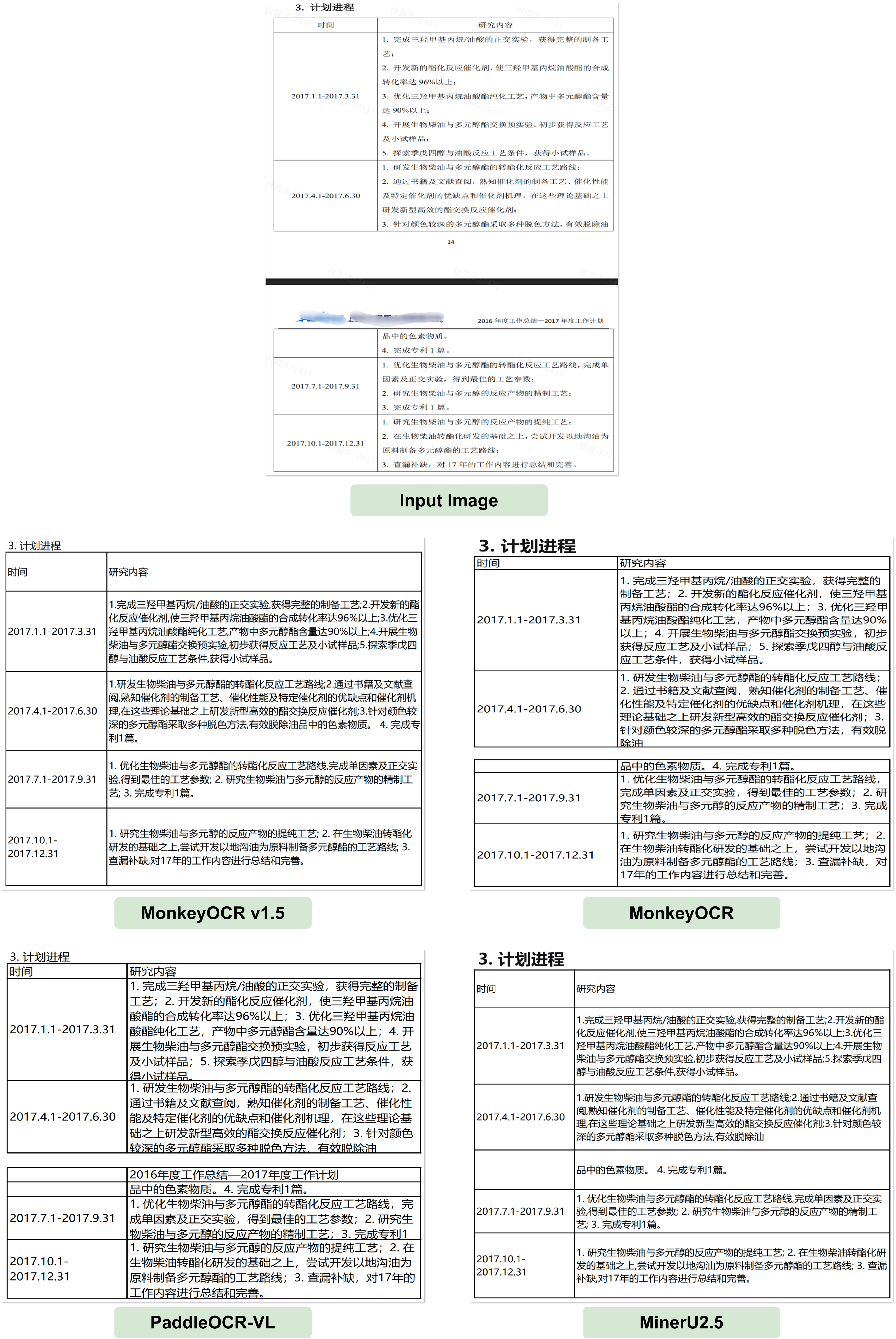}
\caption{\textbf{Cross-Page Table Merging (Example 1)}. Mineru2.5 failed to handle the relationship between the blank cells on the second page and those on the first page, while PaddleOCR-VL was unable to merge cross-page tables. MonkeyOCR v1.5 accurately restored the cross-page table structure.}
\label{fig:Visualization-4}
\end{figure}

\begin{figure}[H]
\centering
\includegraphics[width=0.95\linewidth]{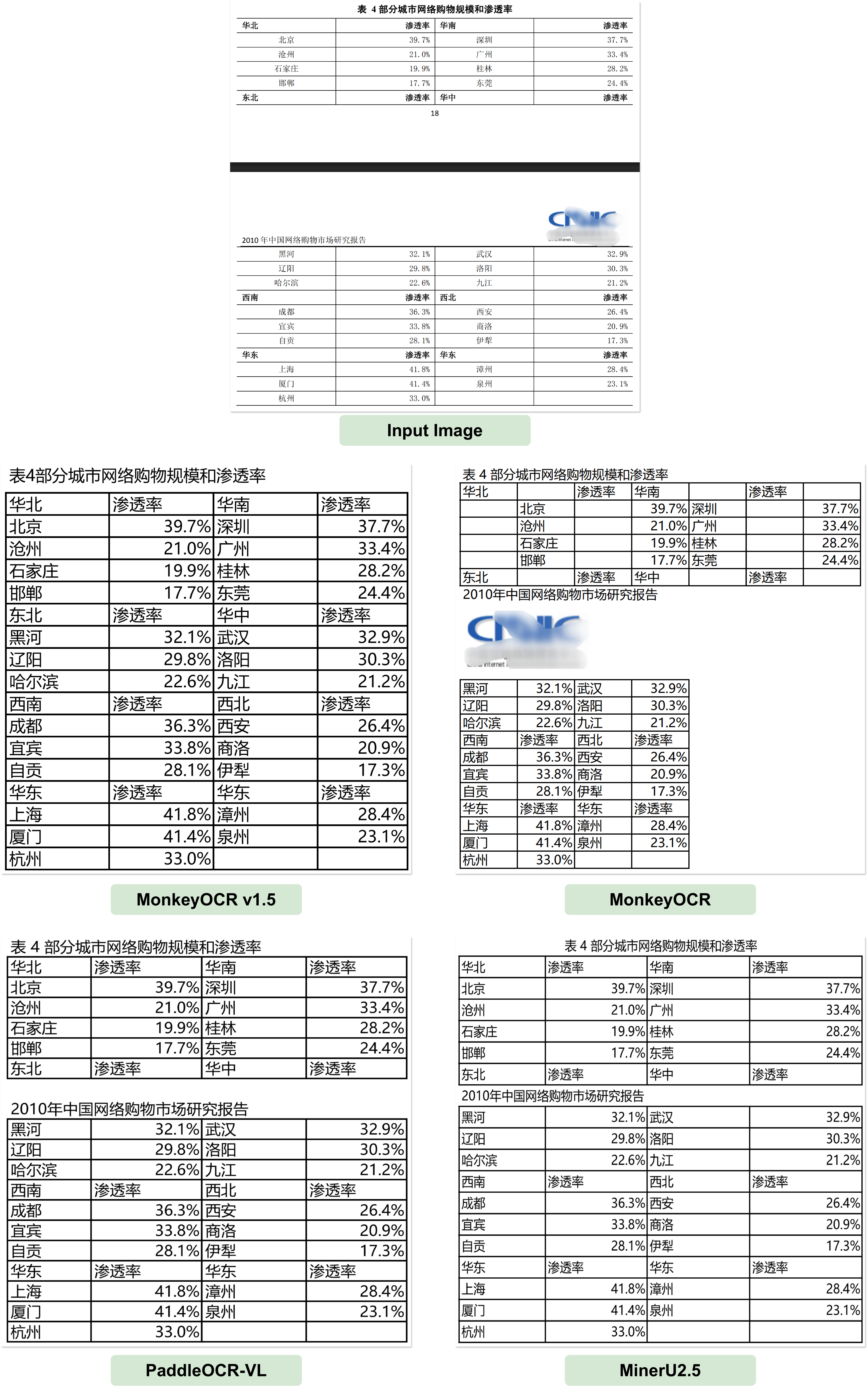}
\caption{\textbf{Cross-Page Table Merging (Example 2)}. Both MinerU2.5 and PaddleOCR-VL failed to restore the complete structure of the cross-page table, as their processing was interrupted by the header between the two pages.}
\label{fig:Visualization-5}
\end{figure}

We present qualitative comparisons for layout analysis, embedded image detection and restoration, and cross-page table merging. Detailed examples are provided in the appendix.
For layout analysis, Fig.~\ref{fig:Visualization-1} shows that our method correctly identifies all images and tables. For embedded image processing, Fig.~\ref{fig:Visualization-2} demonstrates that our pipeline restores both the table structure and the embedded figures. Fig.~\ref{fig:Visualization-3} further confirms complete image restoration by our method.
For cross-page table merging, the primary challenge is preserving structural continuity across page breaks. In Fig.~\ref{fig:Visualization-4}, MonkeyOCR v1.5 reconstructs the full table without structural discontinuities. In Fig.~\ref{fig:Visualization-5}, our method suppresses header interference and retains the full cross-page table.

\begin{figure}[H]
\centering
\includegraphics[width=\linewidth]{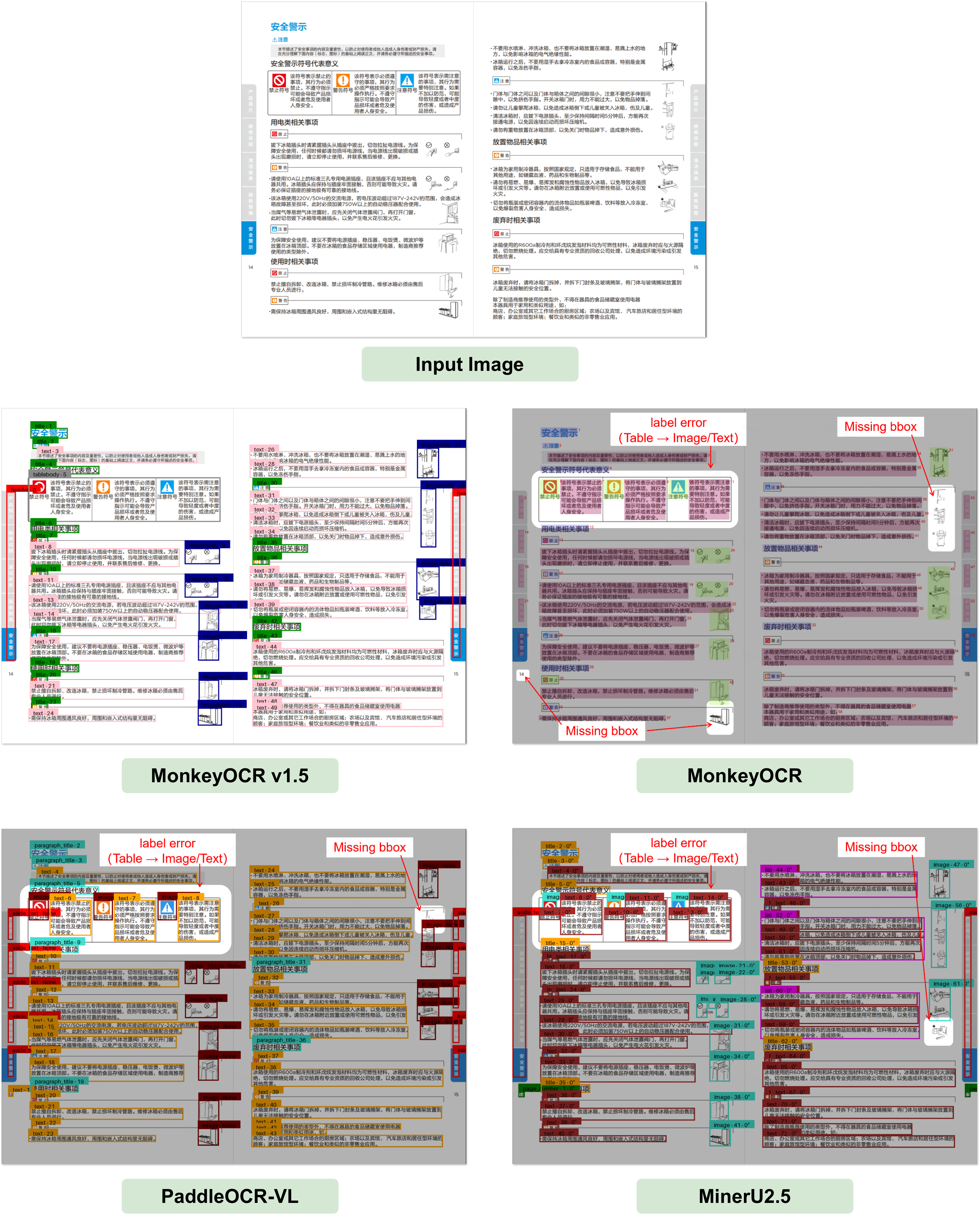}
\caption{\textbf{Layout Analysis Comparison}. Our method accurately identifies all images and tables, while other approaches misclassify table structures as separate text and images.}
\label{fig:Visualization-1}
\end{figure}

\section{Related Work}

\subsection{Traditional pipeline-based methods}
Pipeline-based methods deconstruct the document parsing workflow into a sequence of specialized sub-tasks. These tasks, such as layout analysis~\cite{rt-detr,layoutlmv3,yolov10}, reading order prediction~\cite{layoutreader}, text and formula detection~\cite{easyocr,ppocr,unimernet}, and table structure recognition~\cite{tabformer,tablemaster,docgenome}, are each handled by an independent model. This modular design allows for flexible integration.
For instance, Marker~\cite{marker} supports various document formats by combining modules for OCR, layout analysis, and table recognition. It also leverages Large Language Models (LLMs) to improve cross-page table merging and inline formula parsing. Similarly, systems like MinerU~\cite{mineru} and ppstruct-v3 first use layout models to identify document elements. Subsequently, they employ text and formula detection models to locate the precise positions of text and formulas within these elements, which are then sent to recognition models. A final reading order prediction step assembles all recognized components into a structured output.
Although these pipeline-based approaches benefit from their modularity and have achieved notable success, they are susceptible to error propagation, where inaccuracies in early stages can cascade through the system.

\subsection{LMM-Based Document Parsing Models}

Large Multimodal Models have significantly advanced the field of document understanding. Prior work has improved model capabilities from multiple perspectives: TextMonkey~\cite{monkey,textmonkey} enhances fine-grained visual perception through high-resolution cropping; mPLUG-DocOwl2~\cite{mplug-docowl2} introduces cross-page modeling to enable structural reasoning across multi-page documents; and InternVL3~\cite{internvl} leverages joint pretraining to strengthen cross-modal alignment and long-context understanding. More recently, general-purpose LMMs~\cite{qwen2.5-vl,gemini25,internvl} have been applied to document parsing by processing entire document images in a single pass. MonkeyOCR introduces a triplet paradigm that streamlines pipeline-based methods, reduces compounding errors, and avoids the inefficiencies of full-image processing. MinerU 2.5~\cite{mineru2.5} unifies layout and reading-order prediction within a single multimodal model, while PPOCR-VL~\cite{paddleocrvl} adopts a similar three-stage approach, using lightweight models for structure detection and relation prediction and a large model for content recognition. OCRFlux~\cite{OCRFLUX} excels at page-level parsing, converting PDFs and images into clean Markdown with support for complex layouts, tables, equations, and cross-page merging. Dots.ocr~\cite{dotsocr} offers a compact multilingual solution integrating layout detection and content recognition. Nanonets-OCR2~\cite{Nanonets-OCR2} enables advanced image-to-Markdown conversion and context-aware VQA with accurate structure extraction. OlmOCR 2~\cite{olmocr2} leverages RL-trained 7B VLMs with verifiable unit tests, achieving significant improvements in extracting tables, equations, and multi-column layouts from PDFs.

\section{Conclusion}

In this paper, we presented {MonkeyOCR v1.5}, a unified vision–language document parsing framework that advances both structural understanding and content recognition of complex documents. By jointly predicting layout and reading order through a large multimodal model and performing localized content recognition in a two-stage pipeline, MonkeyOCR v1.5 achieves a balanced trade-off between accuracy and efficiency. The proposed \emph{visual consistency–based reinforcement learning} enables self-supervised refinement of table structures without the need for dense manual annotations, while the \emph{Image-Decoupled Table Parsing (IDTP)} and \emph{Type-Guided Table Merging (TGTM)} modules effectively resolve long-standing challenges such as embedded image handling and cross-page or cross-column table reconstruction. Experimental results on OmniDocBench v1.5 and PubTabNet, and OCRFlux-pubtabnet-single demonstrate that MonkeyOCR v1.5 surpasses previous state-of-the-art methods in both accuracy and robustness. Beyond its superior performance, MonkeyOCR v1.5 is capable of handling challenging scenarios such as embedded table restoration and cross-page table merging, which are difficult for other methods. These capabilities highlight its potential as a foundation model for document understanding in OCRBench~\cite{ocrbench, ocrbenchv2}.

\thispagestyle{plain}
\bibliographystyle{plain}
\bibliography{monkeyocr1.5}

\end{document}